\documentclass{article}
\usepackage{arxiv}
\usepackage[utf8]{inputenc} 
\usepackage[T1]{fontenc}    
\usepackage{hyperref}
\usepackage{ upgreek }
\usepackage{url}            
\usepackage{booktabs}       
\usepackage{amsfonts}       
\usepackage{nicefrac}       
\usepackage{microtype}      
\usepackage{lipsum}
\usepackage{float}
\usepackage{graphicx}
\usepackage{multirow}
\usepackage{array}
\usepackage{comment}
\usepackage{longtable}
\usepackage{ragged2e}
\usepackage{subfig}
\usepackage{amsmath}
\usepackage{longtable}
\usepackage{ amssymb }
\title{Generative Adversarial Networks for Data Augmentation}

\author{
  Angona Biswas   \\
  Research and Development Department, Pioneer Alpha,\\
Dhaka, Bangladesh\\
  \texttt{angonabiswas28@gmail.com} \\
   \And
    MD Abdullah Al Nasim   \\
  Research and Development Department, Pioneer Alpha,\\
Dhaka, Bangladesh\\
  \texttt{nasim.abdullah@ieee.org} \\
   \And
      Al Imran \\
   Research and Development Department, Pioneer Alpha,\\
Dhaka, Bangladesh\\
\texttt{contact.imran00@gmail.com} \\
   \And
Anika Tabassum Sejuty\\
Research and Development Department, Pioneer Alpha,\\
Dhaka, Bangladesh\\
\texttt{sejuty1111@gmail.com}
   \And 
Fabliha Fairooz\\
Research and Development Department, Pioneer Alpha,\\
Dhaka, Bangladesh\\
\texttt{fabilhafairooz@gmail.com}
   \And
  Sai Puppala \\
  Department of Computer Science, University of Alabama at Birmingham\\
Alabama, USA\\
  \texttt{spuppala@uab.edu} \\
   \And
 Sajedul Talukder \\
  Department of Computer Science, University of Alabama at Birmingham\\
Alabama, USA\\
  \texttt{stalukder@uab.edu}} 

\begin{document}
\maketitle

\begin{abstract}
One way to expand the available dataset for training AI models in the medical field is through the use of Generative Adversarial Networks (GANs) for data augmentation. GANs work by employing a generator network to create new data samples that are then assessed by a discriminator network to determine their similarity to real samples. The discriminator network is taught to differentiate between actual and synthetic samples, while the generator system is trained to generate data that closely resemble real ones. The process is repeated until the generator network can produce synthetic data that is indistinguishable from genuine data. GANs have been utilized in medical image analysis for various tasks, including data augmentation, image creation, and domain adaptation. They can generate synthetic samples that can be used to increase the available dataset, especially in cases where obtaining large amounts of genuine data is difficult or unethical. However, it is essential to note that the use of GANs in medical imaging is still an active area of research to ensure that the produced images are of high quality and suitable for use in clinical settings.

\end{abstract}

\section{Introduction}

Data augmentation is an important technique in medical image analysis to improve the robustness and generalization of models. Numerous studies have applied data augmentation techniques, including Generative Adversarial Networks (GANs), to medical images for generating realistic images. A few of the popular examples are described by authors Abdelhalim~\cite{abdelhalim2021data}, and Sun~\cite{sun2020mm} in their conference papers on how Generative Adversarial Network is useful for data augmentation. Abdelhalim~\cite{abdelhalim2021data} in his conference papers explains about on how to create generated pictures of skin lesions using a GAN to enhance the data. On the other side Sun et al.~\cite{sun2020mm} employs Generative Adversarial Networks (GANs) to create artificial medical pictures for data augmentation and shows how well it performs on a task to identify lung nodules. These studies highlight the necessity for rigorous examination of the produced pictures to assure their quality and appropriateness for clinical application, in addition to the potential of GANs for data augmentation in medical image analysis.

\begin{figure}
\centering
\includegraphics[height=5.4cm]{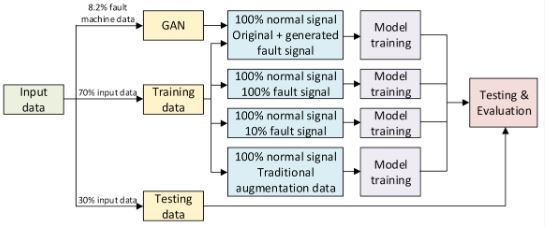}
\caption{Distribution of various dataset types (a) Dataset with the sufficient sample (b) Dataset with a poor sample size. \cite{sampath2021survey}}
\label{gnn1.JPG}
\end{figure}

Figure \ref{gnn1.JPG} is representing the process of augmentation using GANN. In this paper \cite{sampath2021survey},the training and testing procedure is shown in Figure \ref{gnn1.JPG} utilizing both actual data and data produced by the GAN. The input data is split into three groups: training data, which makes up 70\% of the total, testing data, which makes up 30\% of the total, and GAN data, which makes up 11.75\% of the training data and 8.2 percent of the fault machine data.

It is crucial to remember that data augmentation is just one aspect of building robust medical image analysis models, and a thorough evaluation of multiple techniques and models is needed for each specific task and dataset. The most important component of AI applications is data. Lack of sufficient labeled data frequently results in overfitting, which prevents the model from generalizing to new samples. This can be lessened through data augmentation, which effectively increases the volume and diversity of data that the network observes. It is accomplished by applying changes, including such rotation, cropping, shadowing, etc., to an initial dataset in order to artificially create new data. However, figuring out which augmentations will be most effective for the current dataset is not an easy process.

We must understand that data augmentation is just one aspect of building robust medical image analysis models for creating pseudo-realistic images, and extra evaluation of multiple techniques and models might be needed for each specific task requirement and data set. The most important component of AI applications is data. Lack of sufficient labeled data frequently results in overfitting, which prevents the model from generalizing to new samples. This can be lessened through data augmentation, which effectively increases the volume and diversity of data that the network observes. We can accomplish data augmentation by effectively applying needed changes to input images such as rotation, cropping, shadowing, etc., to an original data set for possible artificial data creation which resembles input images. However, figuring out which augmentations will be most effective for the current data set is not an easy process.

For many years, the method of data augmentation has been employed extensively in machine learning and computer vision. By creating additional samples from the existing ones, data augmentation aims to artificially expand the size of the training dataset. This can lessen overfitting and increase the resilience and generalizability of models. This work demonstrates the value of data augmentation in enhancing the performance of deep learning algorithms in computer vision tasks, as demonstrated by numerous recent studies in the field of data augmentation such as Krizhevsky et al.~\cite{krizhevsky2017imagenet}. In their paper, they further explain the data augmentation methods usage like randomized cropping and flipping to expand the ImageNet collection. Since then, data augmentation has become a widely adopted technique in machine learning, with many studies proposing new data augmentation methods and evaluating their performance on various tasks. Some recent research works that emerged in data augmentation are mentioned by Cubuk et al.~\cite{cubuk2019autoaugment}, which explains work on learning enhancement techniques from data. In their paper, they utilized reinforcement learning to automatically learn optimal data augmentation strategies for a specific task. Also, they propose a novel data augmentation technique based on the random mixing of image patches from different samples. In conclusion, data augmentation has a long history in machine learning and computer vision and it is still an active field of study with new approaches and methodologies being put out often. Pseudo images that resemble with original ones are generated by following data augmentation methodologies and we can further enhance the images with the input data by following the below-mentioned approaches:

Flipping: This involves flipping the image horizontally or vertically to generate a new sample.

Rotation: This involves rotating the image by a random angle to generate a new sample.

Scaling: This involves resizing the image to generate a new sample.

Translation: This involves shifting the image by a random offset to generate a new sample.

Cropping: This involves randomly cropping a portion of the image to generate a new sample.

Color augmentation: This involves randomly changing the brightness, contrast, or saturation of the image to generate a new sample.

Mixup: This involves mixing two different images to generate a new sample.

Cutout: This involves randomly masking out a portion of the image to generate a new sample.

These techniques have been widely used in computer vision and machine learning, and be effective in improving the robustness and generalization of models. The different data augmentation techniques in computer vision are evaluated by some of the recent studies including a study by Xiao et al.~\cite{lv2017data}, Cubuk et al.~\cite{cubuk2019autoaugment}, and Xiao et al.~\cite{lv2017data}. The research done by Xiao et al.~\cite{lv2017data} in their paper talks about various data augmentation techniques, including flipping, rotation, and scaling, on image classification tasks. These studies demonstrate the importance of data augmentation in computer vision and machine learning and highlight the need for careful evaluation of different data augmentation techniques for each specific task and dataset.

A novel approach to data augmentation in computer vision has been presented, and it is known as generative adversarial networks (GANs). A generator network plus a discriminator network make up a GAN, a form of deep learning model. The discriminator network is taught to discern between the created samples and the genuine ones, while the generator network is trained to develop new samples that are identical to the original ones. The generation and discriminator are trained in an oppositional fashion, with the generator attempting to make samples that can deceive the discriminator and the discriminator attempting to accurately determine whether a sample is genuine or fabricated. 

One potential advantage of GANs for data augmentation is that they can generate new samples that are diverse and representative of the original dataset. The resilience and applicability of models developed using the additional dataset may benefit from this approach. GANs could be very beneficial for data augmentation in computer vision, and some of the recent studies by Cubuk et al.~\cite{cubuk2019autoaugment}, and Chen et al.~\cite{chen2022generative} in their papers emphasized it. Here, they suggest a GAN-based data augmentation approach for classifying medical images and a Generative adversarial network data augmentation method for segmenting images. Various works highlight the need for more investigation and assessment of GANs in these fields by showcasing the promise of GANs for data augmentation in computer vision and medical imaging analysis.

In \cite{hossainbrain} recent years, both medical image analysis and artificial intelligence have experienced significant growth and development. The development of AI models that can carry out a variety of tasks in computer-aided diagnoses, such as classification tasks, segmentation techniques, image registration, and image synthesis, has been made possible by the increasing accessibility of massive data of medical images and advancements in deep learning algorithms. These AI models have the potential to significantly impact healthcare by providing faster, more accurate, and more cost-effective solutions for medical image analysis. However, there are also important challenges that must be addressed, including the need for high-quality annotated datasets, the development of robust and interpretable models, and the need for careful validation and evaluation of the models. Recent studies which specifically evaluated the performance metrics of AI models in medical image analysis discuss more these challenges. As presented by author Chen et al.~\cite{chen2022generative} in their paper, evaluate the performance of AI models for disease classification and localization on a large dataset of chest X-rays. Also, author Chen et al.~\cite{chen2022generative} presents a deep learning architecture for pulmonary nodule detection in CT images and evaluates its performance on a large dataset of medical images. These studies demonstrate the potential of AI for medical image analysis and highlight the need for continued research and development in this field.

This chapter will cover the lack of medical data and how generative adversarial networks may assist to address it (GANs). 

The shortage of data in the medical field presents a significant obstacle for Artificial Intelligence and also emphasizes the challenges for medical analysis. It can be time-consuming, costly, and susceptible to ethical and legal restrictions to gather high-quality annotated medical photographs. This may result in a lack of data for developing and testing AI models, which might have an impact on their effectiveness and generalizability.

\begin{figure}
\centering
\includegraphics[height=6.2cm]{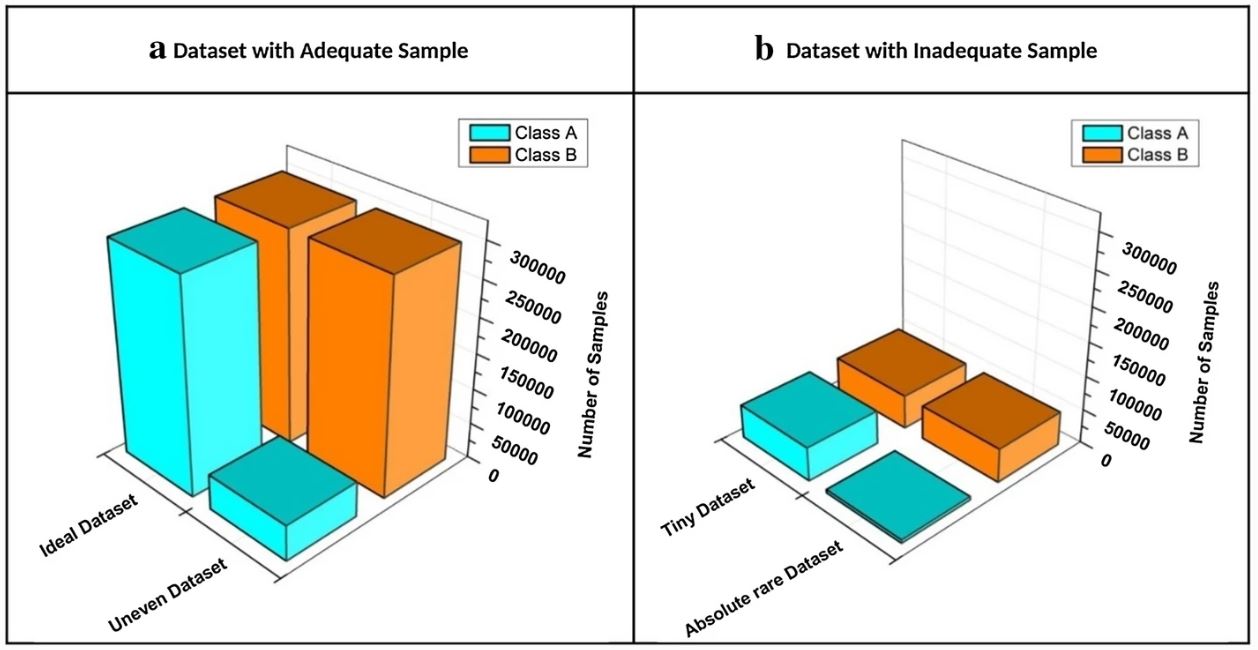}
\caption{Distribution of various dataset types (a) Dataset with a sufficient sample (b) Dataset with a poor sample size~\cite{sampath2021survey}}.
\label{ykh.JPG}
\end{figure}

Several methods have been put out to improve medical data to solve this problem, including generative models like Generational Adversarial Networks (GANs) and Variational Auto-encoders (VAEs). This chapter will discuss the problem of limited data and possible remedies utilizing GANs and other variational auto-encoders (VAEs). These models may be used to create new synthetic pictures that are comparable to the original photos after being trained on a small dataset of medical images. The original dataset may then be supplemented with synthetic pictures, expanding the amount and complexity of the data for developing and testing AI models.

\section{Literature Review}

Medical image analysis and artificial intelligence (AI) have become increasingly important in recent years for improving the accuracy and efficiency of medical diagnosis and treatment~\cite{biswas2022mri}. AI algorithms can be trained on large amounts of medical data to automatically detect and diagnose medical conditions, such as diseases, tumors, and abnormalities, based on medical images. The use of AI in medical image analysis has the potential to revolutionize the healthcare industry by providing more accurate and timely diagnoses, reducing the workload of medical professionals, and improving patient outcomes.

Some examples of AI applications in medical image analysis include computer-aided diagnosis (CAD), image segmentation, image registration, and image synthesis. CAD involves using AI algorithms to assist medical professionals in making diagnoses based on medical images. Image segmentation involves separating an image into different regions or objects of interest. Image registration involves aligning and matching different images of the same patient over time. Image synthesis involves generating new images from existing images for data augmentation, improved visualization, and model training. Several studies have demonstrated the potential of AI in medical image analysis and have shown promising results in various applications.

\subsection {Artificial Intelligence in the Context of Medical Images}

The effective utilization of artificial intelligence (AI) on medical images is crucial for realizing the potential benefits of AI in healthcare, such as improved accuracy and efficiency of medical diagnosis and treatment. To achieve this, several factors must be considered, including data quality, model design, and evaluation methods. Data quality is a key factor in the effective utilization of AI in medical images. Medical images should be of high quality and accurately annotated to ensure that AI algorithms can effectively learn from the data. Additionally, large amounts of medical data are needed to train and evaluate AI models, and the data should be diverse and representative of the population of interest. As presented by author Cai et al.~\cite{cai2020review} in their paper, where they propose a deep learning-based computer-aided diagnosis system and demonstrate its effectiveness on a benchmark dataset. Additionally, Chen et al.~\cite{chen2022generative} in his paper compares the performance of fully trained deep convolutional neural networks (DCNNs) and fine-tuned DCNNs for medical image analysis and demonstrates the effectiveness of fine-tuning for improved performance.

Model design is another important factor in the effective utilization of AI on medical images. AI models should be designed with the specific application in mind, taking into account the type of medical image and the desired output. For example, AI models for medical image segmentation should be designed with the ability to accurately distinguish between different regions or objects of interest in an image. Evaluation methods are also critical for the effective utilization of AI in medical images. For more details about it, we can look into some recent studies where author Cai et al.~\cite{cai2020review} compare the performance of various deep learning algorithms for medical image segmentation and demonstrate the effectiveness of deep learning for this application. In~\cite{hossain2023collaborative} a collaborative federated learning system that enables deep-learning image analysis and classifying diabetic retinopathy without transferring patient data between healthcare organizations has been introduced. The models generated using artificial intelligence should be evaluated on independent datasets to ensure their generalizability and reliability. Additionally, the evaluation metrics used should be relevant to the specific application, we can consider model evaluation metrics such as sensitivity and specificity for medical diagnosis, and the Dice similarity coefficient for medical image segmentation. These studies demonstrate the potential of AI for the effective utilization of medical images in healthcare and highlight the importance of considering data quality, model design, and evaluation methods.

\subsection{Issue of Scarcity of Medical Data}
The scarcity of medical data is a major challenge in the field of healthcare and medical research. Medical data refers to the information generated by medical devices, like imaging devices, and other sources, like electronic health records by author Nasim et al \cite{shah2019brain}. 
  The scarcity of medical data can limit the ability of medical professionals to make accurate diagnoses and develop effective treatments, as well as hinder the development of new medical technologies and research.

The primary reason for data scarcity in the medical field might include data privacy concerns, limited data sharing between institutions, and the cost of data collection and storage. Data privacy concerns make it difficult for medical researchers and companies to access large amounts of medical data for their research and development activities. Limited data sharing between institutions also restricts the availability of medical data, as each institution has its own data collection, storage, and access policies. Finally, the cost of data collection and storage can be prohibitive for many institutions, particularly for those with limited resources. The images available in the medical field are far fewer as compared to other forms of data, such as text and numerical data, for many reasons, some are briefly mentioned below:

Data Privacy Concerns: Medical images often contain sensitive personal information and are subject to strict privacy regulations, such as the Health Insurance Portability and Accountability Act (HIPAA) in the United States. This can make it difficult for researchers and organizations to access large amounts of medical image data \cite{tonmoy2019brain}.

Cost of Data Collection and Storage: Medical imaging, such as MRI and CT scans, require specialized equipment and trained personnel, and the cost of acquiring and storing these images can be high. This limits the number of medical images that can be collected and stored.

Limited Data Sharing between Institutions: Each institution is entitled to its data and they have different policies to access these records. The primary reason for having these policies is to secure medical information as it is very personal and institution likes to keep it confidential. Accessing these data might be very tedious and time-consuming even for researchers.

Annotation Requirements: Medical images often require manual annotation by medical experts, which can be time-consuming and costly. The need for accurate annotations also limits the number of medical images that can be used for training and evaluating machine learning models.

Despite these challenges, efforts are being made to increase the amount of medical image data available, such as data augmentation, synthetic data generation, and federated learning. These approaches aim to overcome the limitations of the scarce amount of medical images and enable the effective utilization of AI in healthcare.

Data augmentation involves generating new data from existing data, while synthetic data generation involves creating artificial data that mimics real-world data. Federated learning involves training machine learning models on multiple institutions' data without actually sharing the data, thereby addressing privacy concerns. Here are a few advancements in federated learning~\cite{talukder2022federated,talukder2022novel,puppala2022towards}.

Despite these efforts, the scarcity of medical data remains a major challenge and requires ongoing research and development to overcome. This is important for improving medical diagnosis and treatment, as well as advancing the field of medical research and technology.

\begin{table}[!h]
\caption{Related work in a nutshell}
\label{T:Ley de Kirchhoff}
\begin{center}
\begin{tabular}  {| p{1.7cm} | p{1.8cm} | p{5.7cm}| p{4 cm} |} 
    \hline
    \textbf{Reference} & \textbf{Publication Year} & \textbf{Workflow} & \textbf{Outcome} \\

    \hline

\cite{chen2022generative} & 2022 & The goal of this study is to synthesize several retinal (or neuronal) pictures with realistic appearances from a tubular structured annotation that is hidden and contains the binary vascular (or neuronal) shape. It was inspired by current developments in generative adversarial networks (GANs) and visual style transfer.  Required 10 training instances.   & The effectiveness of the suggested strategy is supported by extensive experimental evaluations on several retinal fundus and neural imaging applications.  \\ 

      \hline
\cite{8869751} & 2022 & Researchers have enhanced classification by supplementing data using noise-to-image or image-to-image GANs, which may synthesis realistic/diverse extra training pictures to replace the data gap in the true image distribution. Two-step GAN-based DA that creates and improves brain Magnetic Resonance (MR) pictures in order to enhance the DA impact using the GAN combinations.  & Two-step GAN-based DA can greatly outperform the conventional DA alone in tumor identification (raising sensitivity from 93.67\% to 97.48\%).   \\  
      \hline
\cite{radford2015unsupervised} & 2015 & The authors suggest testing the conditional distributions learnt by applying common classification criteria to a conditional version of our model. Using a nearest neighbor classifier to compare actual data to a collection of artificially created conditional samples, we trained a DCGAN using MNIST (splitting off a 10K validation set) and a permutation invariant GAN baseline.  & The DCGAN model successfully models the conditional distributions of this dataset, as evidenced by its test error being equal to that of a closest neighbor classifier fitted on the training dataset. The test error for 10M samples is 1.48\%. \\ 
      \hline
 
\cite{bowles2018gan} & 2018 & The use of GAN-derived synthetic pictures to supplement training data has been examined in this article, and it has been shown to increase performance on two segmentation tasks. The strategy has been demonstrated to perform well in situations with sparse data, whether due to a dearth of accurate data or as a result of class inequality.  & According to a cautious interpretation of the findings from the usual tasks investigated here, supplementing 5–50 labelled picture volumes with an extra 10–100\% GAN-derived synthetic patches has the potential to significantly enhance DSC.\\ 
      \hline   
\end{tabular}
\end{center}
\end{table}

\subsection{Data Augmentation}

Image augmentation is a technique used to artificially increase the size of a dataset by generating new images based on existing ones. The purpose of image augmentation is to reduce overfitting, improve generalization, and increase the robustness of machine learning models when applied to image recognition tasks. Image augmentation is commonly used in computer vision and medical imaging. There are various types of image augmentation techniques, including rotation, scaling, flipping, cropping, and color transformation. These techniques can be applied to images randomly or in a controlled manner to generate new, augmented images. The choice of augmentation techniques depends on the nature of the task and the type of data being used. Image augmentation is very useful for many reasons. Some of the common problems while dealing with deep learning or machine learning algorithms are Overfitting and Underfitting, Overfitting occurs when a machine learning model is too closely fit to the training data and performs poorly on new, unseen data. Image augmentation can help reduce overfitting by generating new, augmented images that can be used to train the model, making it more robust and generalizable. Generating these pseudo images using data augmentation techniques based on existing ones could help improve the generalization of machine learning models to new, unseen data. This can increase the robustness of the models and reduce the risk of poor performance on real-world data. Image augmentation can be used to simulate real-world scenarios, such as changes in lighting conditions, and to test the robustness of machine learning models to these conditions. This can help ensure that the models are suitable for deployment in real-world applications.  Image augmentation can also be used to balance the distribution of classes in a dataset, which can be important for avoiding bias in machine learning models. 

In medical imaging, image augmentation is used to address the scarcity of annotated medical images. Image augmentation is also applied in the clinical imaging field to address the scarcity of annotated medical images. In clinical imaging, it is important to have large datasets of annotated images to train machine-learning models for various tasks, such as disease diagnosis, lesion segmentation, and treatment planning. However, obtaining annotated medical images can be challenging due to ethical and logistical reasons. Image augmentation can also be used to simulate real-world scenarios, such as changes in lighting conditions, and to test the generalization of models to new, unseen data. Overall, image augmentation is a powerful tool for addressing the scarcity of data in computer vision and medical imaging, and has the potential to improve the performance and robustness of machine learning models applied to these fields. 

The use of Generative Adversarial Networks (GANs) for picture-generating tasks is very common. The purpose of picture creation is to create artificial visuals that resemble genuine photos. The generator is instructed to generate synthetic pictures that are identical to genuine images, while the classifier is trained to discriminate between actual and synthetic images.

In the \cite{al2022brain} realm of image generation, GANs have been applied to a variety of tasks, including creating fake pictures of people, things, and environments. These applications have proven that GANs are capable of producing high-quality synthetic pictures that closely resemble genuine photos and are efficient at generating images that are hard to get or are not readily available in huge quantities.

\subsection{Concept of General Adversarial Network}

The role of a Generative Adversarial Network is broadly divided into two parts namely, A generator and a discriminator in deep learning architecture as mentioned by the author Ali et al.~\cite{ali2022role}. The discriminator is taught to distinguish between the synthetic pictures produced by the generator and actual photos from the target dataset, while the generator is trained to produce artificial images that are comparable to a target dataset.

The discriminator seeks to reliably determine whether a picture is genuine or synthetic, while the generator strives to create artificial images that are indistinguishable from actual photos. The generator and discriminator are trained together in a game considering. While the discriminator gets better at telling the difference between actual and fake photos, the generator gets better at creating fake images that seem like real ones with time as presented by author Sauber-Cole et al.~\cite{sauber2022use}

The use of Generational Adversarial Networks (GANs) for picture-generating tasks is very common. The purpose of picture creation is to create artificial visuals that resemble genuine photos. The generator is trained to produce synthetic pictures that are indistinguishable from genuine images, while the discriminator is trained to discriminate between actual and synthetic images. These applications have proven that GANs are capable of producing high-quality synthetic pictures that closely resemble genuine photos and are efficient at generating images that are hard to get or are not readily available in huge quantities. These synthetic images can be used to augment the size of the dataset, improving the accuracy and robustness of machine learning models.

\begin{figure}
\centering
\includegraphics[height=5.4cm]{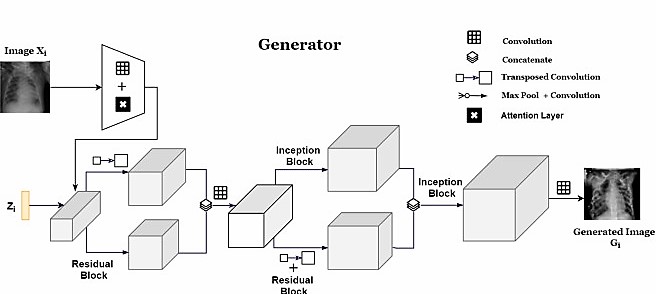}
\caption{Distribution of various dataset types (a) Dataset with the sufficient sample (b) Dataset with a poor sample size.\cite{sampath2021survey}}
\label{gnn22.JPG}
\end{figure}

Yet, conventional data augmentation techniques only yield a small amount of credible alternative data. The performance of CNNs has been enhanced by the use of Generative Adversarial Networks (GANs), which produce fresh data. Yet, compared to CNNs, data augmentation methods for training GANs are less studied. In this study \cite{nasim2021prominence}, authors suggest a novel GAN architecture for augmenting chest X-rays for semi-supervised COVID-19 and pneumonia identification. The suggested GAN may be utilized to efficiently augment data and enhance illness classification accuracy in chest X-rays for pneumonia and COVID-19, according to the authors' research. The construction of the suggested IAGAN's Generator is seen in Fig. \ref{gnn22.JPG}. The generator (G) receives a batch of actual training images and a Gaussian noise vector as input at each cycle. Authors aim to not only use the full image representation using the discriminator, but also to get a lower representation of images fed through the generator for better generalizability of G in generating image data by first encoding the input images using convolution and attention layers to a lower-dimensional representation, before concatenating this representation of the image with the projected noise vector (concatenation happens after goes through a dense layer and non-linearity). The trained generator may employ photos from several classes and produce a wider variety of images thanks to the dual input to the generator.

Once the training process is complete, the generator can be used to generate synthetic images that are similar to the target dataset. These synthetic images can be used to augment the size of the dataset, improving the accuracy and robustness of machine learning models. GANs have been applied to a wide range of applications, including image generation, data augmentation, and semi-supervised learning~\cite{karras2017progressive}. Overall, GANs are a powerful tool for synthesizing new, synthetic images that are similar to real images and are effective for data augmentation in various fields, including medical imaging.

\subsection{GAN concept for medical images}

Generative Adversarial Networks (GANs) is one of the popular deep learning architecture which has been used effectively in several fields, including medical imaging. By creating artificial pictures that closely resemble genuine images, GANs can be employed in medical imaging to supplement the little quantity of annotated clinical data available as stated by author Goodfellow in his book~\cite{goodfellow2020generative}.

A generator and a discriminator are the two primary parts of a GAN elaborately explained in his paper Han et al.~\cite{zhu2017unpaired}. The generator is trained to produce synthetic pictures that are indistinguishable from genuine images, Whereas the discriminator tries to discern the difference between actual and fake pictures, the generator creates synthetic ones. The generator and discriminator are taught in an antagonistic way, with the generator attempting to deceive the discriminator into believing the fake pictures are genuine and the discriminator attempting to accurately identify the true images. GANs can be used to supplement the training data, boosting its variety and lowering the likelihood of overfitting. The employment of artificial pictures may also be utilized to investigate the latent region of the distribution of data and assess how reliable computer vision algorithms are.

GANs have been used for a variety of tasks in medical imaging, including image synthesis, super-resolution, image-to-image translation, and data augmentation. GANs have been used, for instance, to convert MR pictures across different modalities and to create high-resolution MR images with low-resolution MR images \cite{al2022brain}. It is significant to remember that both the richness of the training data and the intricacy of the generator structure affect the accuracy and realism of the synthetic pictures produced by GANs as discussed by Radford in his paper about GAN with unsupervised representation learning~\cite{radford2015unsupervised}. Additionally, GANs are heavily data-driven and may not adapt well to additional data distributions, particularly in medical imaging where the data can be quite diverse and sensitive to inter-patient variability.

\section{Methodology}

A significant obstacle in the realm of medical image analysis is the dearth of medical data. The purpose of this study is to investigate the usage of variational auto-encoders (VAEs) and generative adversarial networks (Generative adversarial) to improve medical data~\cite{kingma2013auto}. The literature on the application of GANs and VAEs for data augmentation will be thoroughly reviewed. This review will focus on the use of these methods in the area of medical image analysis and will cover recent research publications, technical reports, and conference proceedings.

\subsection{Data Collection}
For experimentation, the project will employ a publicly accessible medical imaging collection, such as the MRI, CT Scan, or Chest X-Ray dataset. One portion of the dataset will be utilized for training and the other portion for testing. Data collection is a primary and crucial step before we train our deep learning model using generative adversarial networks (GANs). The GANs' effectiveness and capacity to produce plausible synthetic data can be considerably impacted by the caliber and diversity of the data used to train them, according to the author, Goodfellow et al.~\cite{goodfellow2020generative}. The method of gathering data for GANs involves numerous steps. The selection of the data is also crucial. Choosing the pertinent data for the task at hand is the first step. A publicly accessible medical images collection, like the MNIST or the Chest X-Ray dataset, might be employed in the case of healthcare image analysis. Making sure that the information gathered is diverse and appropriate for the task at hand is crucial. The GANs will be better able to generalize and provide high-quality synthetic data as a result.

\subsection{MRI Pre-processing} 
Once the data has been selected, it needs to be preprocessed to prepare it for training. This may involve resizing the images to a standard size, normalizing the pixel values, and removing any irrelevant information from the images. Image preprocessing is an important step before training Generative Adversarial Networks (GANs) as it can significantly impact the performance and ability of the GANs to generate realistic synthetic data by referred by Karras in his paper about analyzing and improving image quality~\cite{karras2020analyzing}.

\subsection{Workflow of GAN for Data Augmentation}
To implement the GANs and VAEs, deep learning frameworks like TensorFlow or PyTorch will be used. The training dataset will be used to train the GANs and VAEs, and the testing dataset will be used to assess their performance. The preprocessed data may be used with data augmentation techniques to add variation to the training data and to guard against overfitting. Random picture flips, translations, and rotations are common data augmentation techniques.

A training set as well as a validation set should be created from the preprocessed and enhanced data. The validation set is used to assess how well the GANs performed while they were being trained, whereas the training set is used to train the GANs.

\begin{figure}
\centering
\includegraphics[height= 5.00cm]{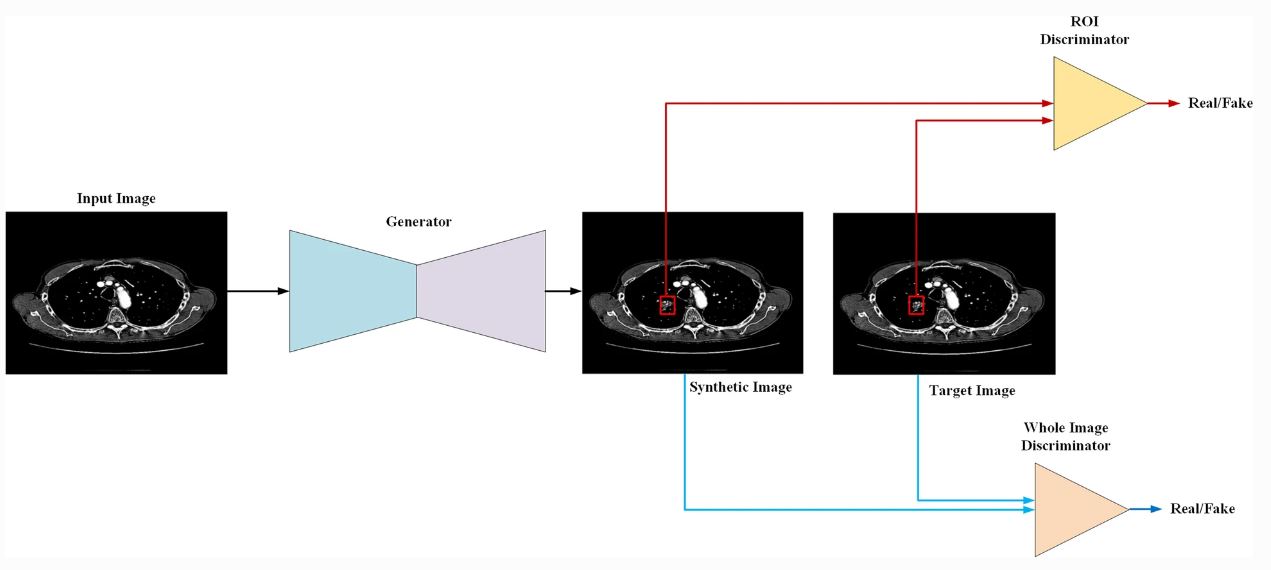}
\caption{The model training process pipeline. According to the supplied picture, the generator first creates ground glass nodules from the backdrop. Second, to determine if the synthetic picture is real or not, either region of interest (ROI) discriminative model (red line) and the entire image discriminator (blue line) extract the features from the ROI and complete image, respectively.~\cite{wang2022generation}}
\label{gg1.JPG}
\end{figure}

\begin{figure}
\centering
\includegraphics[height= 5.00cm]{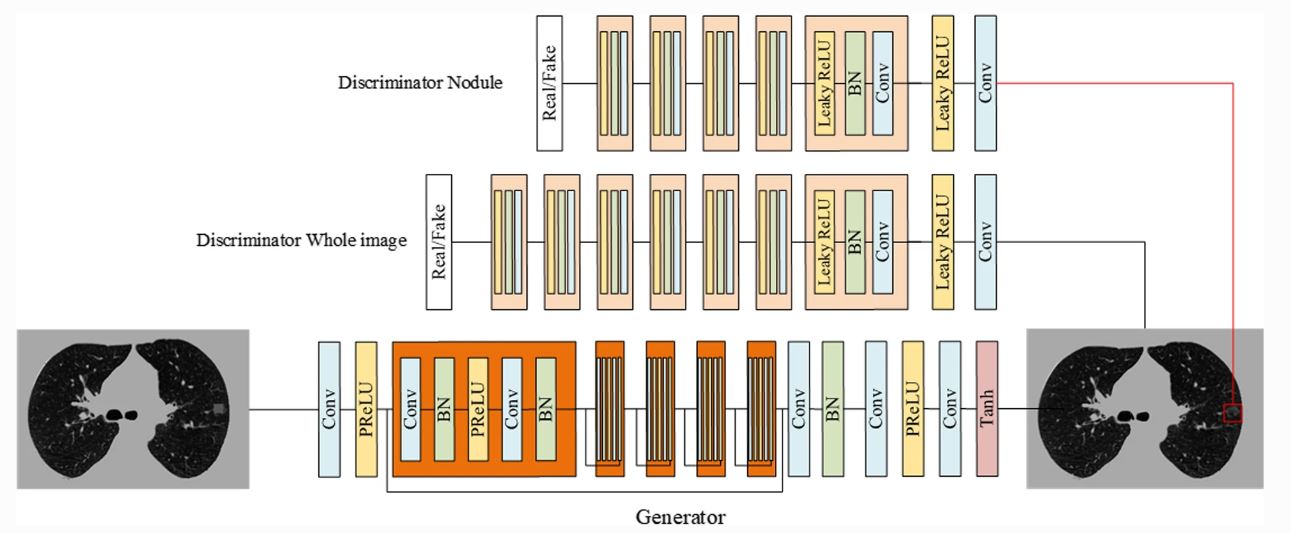}
\caption{The network's structure. In the input mask's place, the generator produces a synthetic ground glass nodule. The batch normalization, the "parametric rectified linear unit" (PReLU) activation function, and convolutional layers with a 3 3 kernel size make up the generator. The batch normalization, the leaky PReLU activation function, and the 3 3 kernel size convolutional layers made up the discriminator.~\cite{wang2022generation}}
\label{gg2.JPG}
\end{figure}

\subsubsection{General Algorithm.}
A generator and a discriminator are the two primary parts of the deep learning technique known as generative adversarial networks (GANs). To create fake data that mimics genuine data, these elements are trained in an adversarial way. 

The basic steps involved in a GAN algorithm are:

1.  Data Preparation: The training data is collected and preprocessed.

2. Generator: The generator is a neural network that maps a random noise vector to a synthetic data sample. The generator is trained to generate synthetic data that resembles real data.

3. Discriminator: The discriminator is a neural network that distinguishes between real and synthetic data. The discriminator is trained to correctly identify real data and to reject synthetic data.

4. Adversarial Training: The generator and the discriminator are trained simultaneously, in an adversarial manner. The generator tries to generate synthetic data that the discriminator cannot distinguish from the real data, while the discriminator tries to correctly identify the real data and reject the synthetic data generated by the generator.

5. Synthetic Data Generation: After the training process is completed, the generator can be used to generate synthetic data that resembles real data.

The quality of the synthetic data generated by a GAN depends on the complexity of the generator and discriminator, the quality of the training data, and the number of iterations performed during the training process.

\subsection{The Process of Data Augmentation Using GAN}
Generative Adversarial Networks (GANs) perform data augmentation by synthesizing new, artificial data samples that resemble the real data~\cite{zhang2020insufficient}. The GAN consists of two main components: a generator and a discriminator. The generator is a neural network that maps a random noise vector to a synthetic data sample.

The discriminator is a neural network that distinguishes between real and synthetic data. The discriminator is trained to correctly identify real data and to reject synthetic data. The generator and the discriminator are trained simultaneously, in an adversarial manner. The generator tries to generate synthetic data that the discriminator cannot distinguish from the real data, while the discriminator tries to correctly identify the real data and reject the synthetic data generated by the generator as presented by author Bowles et al.~\cite{bowles2018gan}. After the training process is completed, the generator can be used to generate synthetic data that resembles real data. This synthetic data can be used to augment the real data, increasing the size of the training data set, and improving the performance of machine learning algorithms trained on this data.

\subsection{Data Augmentation Using Variational Auto-Encoders (VAEs)}
Variational Autoencoders (VAEs) can be used for data augmentation by generating synthetic data samples from the same underlying distribution as the real data. VAEs consist of an encoder network that maps the input data to a lower-dimensional representation, and a decoder network that maps the lower-dimensional representation back to the original data space~\cite{saldanha2022data}. The encoder and decoder are trained together in an unsupervised manner, such that the decoder can reconstruct the original data from the lower-dimensional representation.

\begin{figure}
\centering
\includegraphics[height= 4.00cm]{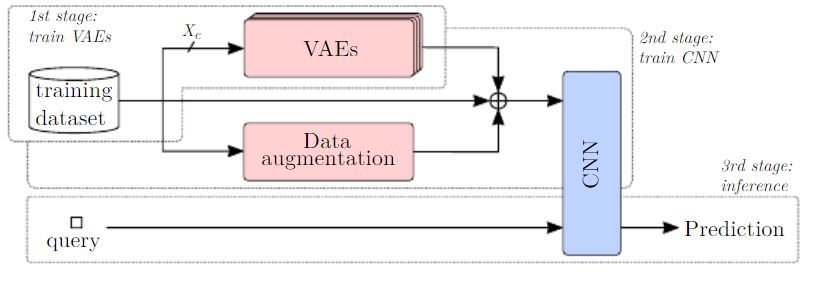}
\caption{General description of the suggested technique. ~\cite{garay2019data}}
\label{vae.JPG}
\end{figure} 

To use VAEs for data augmentation, the encoder-decoder architecture can be used to generate synthetic data samples as presented by author Garay-Maestre,~\cite{garay2019data}. This can be done by sampling from the prior distribution over the lower-dimensional representation, and then passing this sample through the decoder network to obtain a synthetic data sample in the original data space.

The one advantage of VAE as compared to that of GAN for data augmentation is that they can generate synthetic data that is similar to the real data in terms of structure and distribution as explained by author Fuertes,~\cite{fuertes2020variational}. This can help improve the performance of machine learning algorithms by increasing the size of the training data set and helping to avoid overfitting.

\subsection{Result Analysis}

The results of the study will be analyzed and discussed in terms of the performance of the GANs and VAEs for data augmentation. The results will be compared to existing approaches for data augmentation, and the advantages and disadvantages of the GANs and VAEs will be discussed. 
The study will conclude with a summary of the findings and a discussion of the implications of the results for the field of medical image analysis. Recommendations for future research in this area will also be provided. 

Generative Adversarial Networks (GANs) are effective for image augmentation because they can generate synthetic data that resembles real data. This synthetic data can be used to augment the real data, increasing the size of the training data set and improving the performance of machine learning algorithms trained on this data.

GANs can generate high-quality synthetic data because they are trained on real data. The generator network maps a random noise vector to a synthetic data sample, while the discriminator network is trained to distinguish between real and synthetic data. The generator and discriminator are trained simultaneously in an adversarial manner, with the generator trying to generate synthetic data that the discriminator cannot distinguish from real data, and the discriminator tries to correctly identify real data and reject synthetic data.

This adversarial training process results in the generator learning to generate synthetic data that closely resembles the real data, and the discriminator learning to correctly identify real data. As a result, the synthetic data generated by the GAN can be used to augment the real data and increase the size of the training data set, leading to improved performance of machine learning algorithms.

\section{Conclusion}

Generative Adversarial Networks (GANs) may be utilized for data augmentation for creating synthetic samples to increase the size of the training dataset. In this procedure, new data samples are generated by a generator network and then assessed by a discriminator network to see whether they are sufficiently comparable to the original samples. While the discriminator network is taught to discriminate between actual and created samples, the generator network is trained to increase its capacity to generate data that are comparable to real ones. The enhanced dataset may then be used to train a model for the desired objective. The procedure is repeated until the generator network can generate synthetic samples that are indistinguishable from genuine ones. Medical image analysis has used Generative Adversarial Networks (GANs) for a variety of tasks, including data augmentation, picture creation, and domain adaptation. In the field of medical imaging, GANs can create synthetic samples that can be utilized to increase the dataset that is already accessible, especially when collecting significant volumes of actual data is challenging or morally problematic. It is important to emphasize that research into the application of GANs in radiography is still ongoing, and a thorough evaluation of the produced pictures is necessary to assure their quality and appropriateness for clinical applications.
\bibliographystyle{plain}
\bibliography{main.bib}

\end{document}